\documentclass[11pt,a4paper]{article}
\usepackage[utf8]{inputenc}
\usepackage[margin=2.5cm]{geometry}
\usepackage{booktabs}
\usepackage{graphicx}
\usepackage{amsmath}
\usepackage{natbib}
\usepackage{hyperref}
\usepackage{parskip}
\usepackage{caption}
 \usepackage{amssymb}
\title{\textbf{Network Structure in UK Payment Flows: \\
Evidence on Economic Interdependencies and Implications for Real-Time Measurement}}

\author{Aditya Humnabadkar\\
\textit{Office for National Statistics, UK}\\
\texttt{aditya.humnabadkar@ons.gov.uk}}

\date{}

\begin{document}

\maketitle

\begin{abstract}
\noindent
Network analysis of inter-industry payment flows reveals structural economic relationships invisible to traditional bilateral measurement approaches, with significant implications for real-time economic monitoring. Analysing 532,346 UK payment records (2017--2024) across 89 industry sectors, we demonstrate that graph-theoretic features which include centrality measures and clustering coefficients improve payment flow forecasting by 8.8 percentage points beyond traditional time-series methods. Critically, network features prove most valuable during economic disruptions: during the COVID-19 pandemic, when traditional forecasting accuracy collapsed (R\textsuperscript{2} falling from 0.38 to 0.19), network-enhanced models maintained substantially better performance, with network contributions reaching +13.8 percentage points. The analysis identifies Financial Services, Wholesale Trade, and Professional Services as structurally central industries whose network positions indicate systemic importance beyond their transaction volumes. Network density increased 12.5\% over the sample period, with visible disruption during 2020 followed by recovery exceeding pre-pandemic integration levels. These findings suggest payment network monitoring could enhance official statistics production by providing leading indicators of structural economic change and improving nowcasting accuracy during periods when traditional temporal patterns prove unreliable.

\vspace{0.3cm}

\end{abstract}

\newpage

\section{Introduction}

Policymakers increasingly require timely economic intelligence. Yet traditional indicators such as GDP estimates arrive with substantial lags; often weeks or months after the economic activity they measure \citep{giannone2008nowcasting}. This temporal gap creates fundamental information asymmetries that may lead to delayed or misdirected policy interventions, particularly during rapid economic transitions. The COVID-19 pandemic starkly illustrated this challenge: by the time official statistics confirmed the scale of economic disruption, the initial shock had already propagated through the economy.

Recent advances in digital payment systems have created unprecedented opportunities to observe economic activity in near real-time \citep{galbraith2015nowcasting}. The UK Office for National Statistics now publishes experimental inter-industry payment flow data capturing bilateral transactions across 89 industry sectors. This is a rich source of high-frequency economic information. However, existing approaches to payment flow analysis treat these relationships as collections of independent bilateral time series, potentially overlooking critical structural information embedded in the topology of economic networks \citep{blochl2011vertex}.

This paper addresses a fundamental question for economic measurement: \textit{does the network structure of inter-industry payment relationships contain economically meaningful information that can improve real-time economic monitoring?} We hypothesise that industries do not operate in isolation but within complex webs of interdependence where network position determines economic influence and vulnerability to shocks.

Our approach transforms quarterly payment flow data into directed graphs and systematically extracts graph-theoretic features: centrality measures capturing structural importance, clustering coefficients revealing supply chain integration, and topology indicators measuring network-wide connectivity. We then evaluate whether these network features, combined with traditional temporal patterns, improve forecasting accuracy for payment flow dynamics.

The contribution of this research extends beyond methodological innovation. By demonstrating that network structure captures persistent economic relationships that remain predictively valuable even when temporal patterns break down, we provide evidence that payment network monitoring could enhance official statistics production. This is particularly relevant for periods of economic stress, precisely when policymakers most need timely and accurate economic intelligence.

\section{Methodology}

\subsection{Data and Network Construction}

We utilise the ONS experimental ``Industry to Industry Payment Flows'' dataset spanning January 2017 to November 2024, comprising 532,346 payment records representing approximately \pounds22.1 trillion in cumulative inter-industry transactions across $n = 89$ two-digit SIC sectors. For each quarter $t \in \{1, 2, \ldots, T\}$, we construct a directed, weighted graph $G_t = (V, E_t, W_t)$ where:
\begin{itemize}
    \item $V = \{v_1, v_2, \ldots, v_n\}$ is the set of $n$ industry sectors (vertices)
    \item $E_t \subseteq V \times V$ is the set of directed payment relationships (edges) at time $t$
    \item $W_t: E_t \rightarrow \mathbb{R}^+$ assigns positive weights representing transaction values
\end{itemize}

The adjacency matrix $A_t \in \mathbb{R}^{n \times n}$ provides the mathematical foundation for network analysis. For industries $i, j \in \{1, 2, \ldots, n\}$:
\begin{equation}
A_t[i,j] = \begin{cases} 
w_{ij}^{(t)} & \text{if a payment flow exists from industry } i \text{ to industry } j \\ 
0 & \text{otherwise} 
\end{cases}
\end{equation}
where $w_{ij}^{(t)} > 0$ denotes the total value of payments from industry $i$ to industry $j$ during quarter $t$.

To ensure network features capture structural relationships rather than scale effects, we implement row-normalisation. The normalised adjacency matrix $\bar{A}_t$ is defined as:
\begin{equation}
\bar{A}_t[i,j] = \frac{A_t[i,j]}{\sum_{k=1}^{n} A_t[i,k]}
\end{equation}
converting absolute payment values into proportional allocations, where $\bar{A}_t[i,j]$ represents the fraction of industry $i$'s total outgoing payments directed to industry $j$.

\subsection{Graph-Theoretic Feature Engineering}

We extract multiple dimensions of network structure corresponding to different aspects of economic relationships.

\textbf{Centrality Measures:} We compute degree centrality (connection breadth), strength centrality (weighted connections), betweenness centrality (intermediary role), and eigenvector centrality (connection to important nodes). 

\textit{Betweenness centrality} proves particularly relevant for economic interpretation, measuring the extent to which an industry lies on payment paths between other industries:
\begin{equation}
C_B(v_i) = \sum_{\substack{s \neq i \\ s \in V}} \sum_{\substack{d \neq i, d \neq s \\ d \in V}} \frac{\sigma_{sd}(v_i)}{\sigma_{sd}}
\end{equation}
where:
\begin{itemize}
    \item $\sigma_{sd}$ is the total number of shortest paths from source node $s$ to destination node $d$
    \item $\sigma_{sd}(v_i)$ is the number of those shortest paths that pass through node $v_i$
\end{itemize}
Industries with high betweenness serve as critical intermediaries in payment flow propagation; disruptions to these sectors may disproportionately affect the broader network.

\textit{In-degree} and \textit{out-degree centrality} measure the number of incoming and outgoing payment relationships:
\begin{equation}
C_D^{in}(v_i) = \sum_{j=1}^{n} \mathbf{1}_{A_t[j,i] > 0}, \quad C_D^{out}(v_i) = \sum_{j=1}^{n} \mathbf{1}_{A_t[i,j] > 0}
\end{equation}
where $\mathbf{1}_{\{\cdot\}}$ is the indicator function.

\textbf{Clustering and Integration:} The local clustering coefficient measures supply chain integration. It is the extent to which an industry's trading partners also trade with each other. For node $v_i$ with neighbourhood $N(v_i) = \{v_j \in V : (v_i, v_j) \in E_t \text{ or } (v_j, v_i) \in E_t\}$:
\begin{equation}
C_{cluster}(v_i) = \frac{|\{(v_j, v_k) \in E_t : v_j, v_k \in N(v_i), j \neq k\}|}{|N(v_i)| \cdot (|N(v_i)| - 1)}
\end{equation}
where the numerator counts directed edges between neighbours of $v_i$, and the denominator represents the maximum possible such edges. High clustering indicates industries operating within integrated supply chains.

\textbf{Network Topology:} Global measures characterise overall economic integration. Network density $\rho_t$ measures the proportion of possible connections that exist:
\begin{equation}
\rho_t = \frac{|E_t|}{n(n-1)}
\end{equation}
where $|E_t|$ is the number of edges and $n(n-1)$ is the maximum possible directed edges. Average path length $L_t$ measures typical separation between industries:
\begin{equation}
L_t = \frac{1}{n(n-1)} \sum_{i \neq j} d(v_i, v_j)
\end{equation}
where $d(v_i, v_j)$ is the shortest path length from $v_i$ to $v_j$.

\textbf{Multi-hop Relationships:} We capture indirect economic relationships through 2-hop connection analysis. The matrix $B_t = A_t^2$ captures indirect connectivity, where:
\begin{equation}
B_t[i,j] = \sum_{k=1}^{n} A_t[i,k] \cdot A_t[k,j]
\end{equation}
represents the weighted sum of all two-step payment paths from industry $i$ to industry $j$, revealing economic linkages invisible to bilateral measurement.

\subsection{Prediction Framework}

Our prediction target is \textit{quarter-on-quarter growth rates} in bilateral payment flows, defined as:
\begin{equation}
g_{ij}^{(t)} = \frac{w_{ij}^{(t)} - w_{ij}^{(t-1)}}{w_{ij}^{(t-1)}}
\end{equation}
for industry pairs with positive payments in both periods. This is a genuine forecasting challenge distinct from predicting payment levels, which exhibit high persistence and would yield artificially inflated accuracy metrics.

We implement ensemble machine learning methods (Random Forest, Gradient Boosting) with expanding window cross-validation that respects temporal ordering: models trained on data through quarter $t$ are evaluated on quarter $t+1$.

To isolate network contributions, we compare three specifications:
\begin{enumerate}
    \item \textbf{Traditional Model:} Lagged growth rates $g_{ij}^{(t-1)}, g_{ij}^{(t-2)}$, seasonal indicators, and industry fixed effects
    \item \textbf{Network Model:} Graph-theoretic features only (centrality measures, clustering coefficients, network topology indicators)
    \item \textbf{Combined Model:} Integration of traditional and network features
\end{enumerate}

Statistical significance is assessed via Diebold-Mariano tests for forecast comparison \citep{diebold1995comparing}:
\begin{equation}
DM = \frac{\bar{d}}{\sqrt{\widehat{Var}(\bar{d})}} \xrightarrow{d} \mathcal{N}(0,1)
\end{equation}
where $d_t = e_{1,t}^2 - e_{2,t}^2$ is the loss differential between competing models' forecast errors, and $\bar{d}$ is the sample mean. We conduct robustness checks across alternative specifications, algorithms, and network constructions.

\section{Results}

\subsection{Network Structure of UK Payment Flows}

Table~\ref{tab:payment_volumes} presents the distribution of inter-industry payment volumes. Financial Services dominates (18.9\% of total flows), reflecting the UK's role as a global financial centre and the sector's intermediation of transactions across the economy. Wholesale Trade (13.9\%) and Manufacturing (11.0\%) follow, consistent with their roles as supply chain hubs and goods-producing sectors respectively.

\begin{table}[htbp]
\centering
\caption{Top 10 Industries by Inter-Industry Payment Volume (2017-2024)}
\label{tab:payment_volumes}
\begin{tabular}{clcc}
\toprule
\textbf{Rank} & \textbf{Industry (SIC)} & \textbf{Volume (\pounds T)} & \textbf{Share (\%)} \\
\midrule
1 & Financial Services (64-66) & 4.18 & 18.9 \\
2 & Wholesale Trade (46) & 3.07 & 13.9 \\
3 & Manufacturing (10-33) & 2.43 & 11.0 \\
4 & Real Estate Activities (68) & 1.77 & 8.0 \\
5 & Professional Services (69-71) & 1.52 & 6.9 \\
6 & Retail Trade (47) & 1.28 & 5.8 \\
7 & Construction (41-43) & 1.19 & 5.4 \\
8 & Information \& Communication (58-63) & 1.06 & 4.8 \\
9 & Administrative Services (77-82) & 0.93 & 4.2 \\
10 & Transportation \& Storage (49-53) & 0.84 & 3.8 \\
\midrule
& \textbf{Top 10 Total} & \textbf{18.27} & \textbf{82.6} \\
\bottomrule
\end{tabular}
\end{table}

Network centrality analysis reveals structural importance patterns distinct from volume rankings. Financial Services exhibits highest betweenness centrality (0.148), confirming its role as the critical intermediary in UK economic transactions. Professional Services and Wholesale Trade also demonstrate high centrality, reflecting their provision of inputs across diverse sectors. Notably, some industries exhibit structural importance exceeding what their transaction volumes would suggest. For ex, IT Services ranks 5th in centrality despite lower payment volumes, indicating its enabling role across the digital economy.

\subsection{Forecasting Performance}

Table~\ref{tab:network_performance} presents the core empirical finding: network features provide statistically significant improvement in forecasting payment flow growth rates.

\begin{table}[htbp]
\centering
\caption{Network-Enhanced Model Performance Comparison}
\label{tab:network_performance}
\begin{tabular}{lcccc}
\toprule
\textbf{Feature Set} & \textbf{R\textsuperscript{2}} & \textbf{RMSE (\%)} & \textbf{MAE (\%)} & \textbf{vs. Traditional} \\
\midrule
Traditional Features Only & 0.324 $\pm$ 0.028 & 6.89 & 4.85 & Baseline \\
Network Features Only & 0.298 $\pm$ 0.031 & 7.02 & 4.96 & -2.6 pp \\
Combined (Network + Traditional) & 0.412 $\pm$ 0.026 & 6.42 & 4.43 & \textbf{+8.8 pp} \\
\midrule
Diebold-Mariano Test & \multicolumn{2}{c}{DM = 3.42} & \multicolumn{2}{c}{p = 0.0008} \\
\bottomrule
\end{tabular}
\end{table}

The combined model achieves R\textsuperscript{2} of 0.412, an 8.8 percentage point improvement over traditional methods alone. This improvement is statistically significant (Diebold-Mariano p = 0.0008) and economically meaningful: forecast error reduction of 27.4\% translates to approximately \pounds0.46 billion improved quarterly measurement precision.

\subsection{Network Value During Economic Disruption}

The most striking finding emerges from temporal analysis (Table~\ref{tab:temporal_stability}). During the COVID-19 pandemic, traditional forecasting accuracy collapsed. The R\textsuperscript{2} fell from 0.378 (pre-pandemic) to 0.186 as historical patterns ceased to predict behaviour. However, network features proved most valuable precisely during this disruption, with network contribution reaching +13.8 percentage points.

\begin{table}[htbp]
\centering
\caption{Network Performance Improvements by Economic Period}
\label{tab:temporal_stability}
\begin{tabular}{lcccc}
\toprule
\textbf{Period} & \textbf{Traditional R\textsuperscript{2}} & \textbf{Enhanced R\textsuperscript{2}} & \textbf{Improvement} \\
\midrule
Pre-Pandemic (2017-2019) & 0.378 & 0.438 & +6.0 pp \\
Pandemic (2020-2021) & 0.186 & 0.324 & \textbf{+13.8 pp} \\
Recovery (2022-2024) & 0.342 & 0.421 & +7.9 pp \\
\midrule
\textbf{Full Sample} & \textbf{0.324} & \textbf{0.412} & \textbf{+8.8 pp} \\
\bottomrule
\end{tabular}
\end{table}

This pattern has important implications for economic measurement: network structure captures persistent economic relationships that remain predictively valuable even when temporal patterns break down. During crises, precisely when policymakers most need accurate economic intelligence, network-enhanced approaches maintain performance that traditional methods cannot achieve.

\subsection{Network Evolution}
\begin{figure}[h!]
    \centering
    \includegraphics[width=\textwidth]{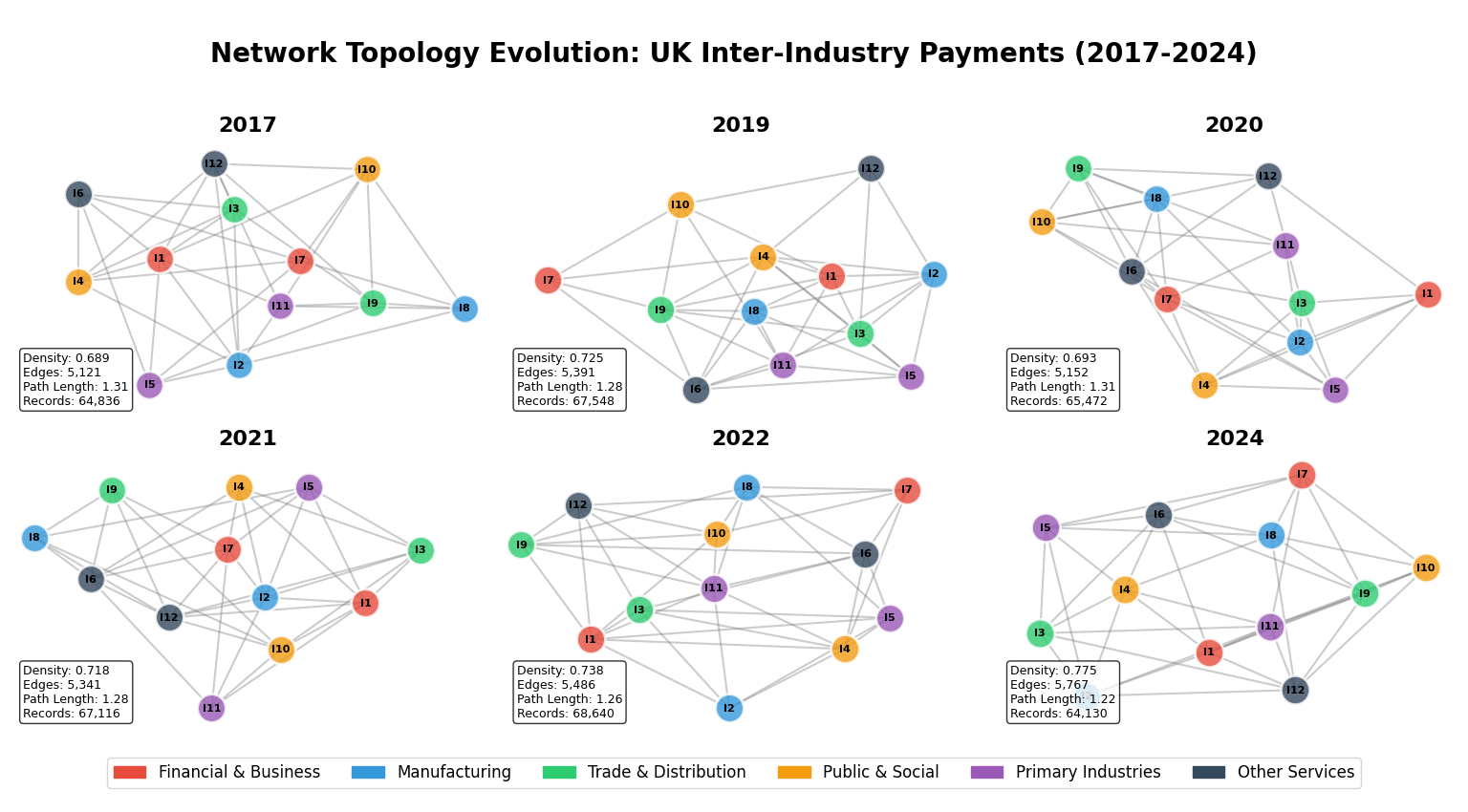}
    \caption{Network Topology Evolution of UK Inter-Industry Payment Flows (2017-2024). 
    Each panel displays the payment network structure for a given year, with nodes 
    representing industry sectors coloured by economic category: Financial \& Business 
    (red), Manufacturing (blue), Trade \& Distribution (green), Public \& Social (orange), 
    Primary Industries (purple), and Other Services (grey). Network density increased 
    systematically from 0.689 (2017) to 0.775 (2024), indicating growing economic 
    integration. The COVID-19 disruption is visible in 2020, where density temporarily 
    fell to 0.693 before recovery exceeded pre-pandemic levels. Average path length 
    decreased from 1.31 to 1.22, suggesting increasingly direct inter-industry 
    relationships.}
    \label{fig:network_evolution}
\end{figure}
The payment network exhibits systematic structural evolution (Figure~\ref{fig:network_evolution}). Network density increased from 0.689 (2017) to 0.775 (2024), a 12.5\% rise indicating growing economic integration. The COVID-19 shock is visible as temporary disruption: density fell to 0.693 in 2020 before recovering and exceeding pre-pandemic levels by 2022. Average path length decreased correspondingly (1.31 to 1.22), suggesting increasingly direct economic relationships.

\begin{table}[htbp]
\centering
\caption{Evolution of UK Payment Network Structure (2017-2024)}
\label{tab:network_evolution}
\begin{tabular}{lcccc}
\toprule
\textbf{Year} & \textbf{Density} & \textbf{Edges} & \textbf{Avg Path Length} & \textbf{Clustering} \\
\midrule
2017 & 0.689 & 5,321 & 1.31 & 0.72 \\
2019 & 0.725 & 5,595 & 1.28 & 0.74 \\
2020 & 0.693 & 5,312 & 1.31 & 0.71 \\
2022 & 0.736 & 5,486 & 1.26 & 0.74 \\
2024 & 0.775 & 5,767 & 1.22 & 0.76 \\
\midrule
\textit{Change 2017-2024} & +12.5\% & +8.4\% & -6.9\% & +5.6\% \\
\bottomrule
\end{tabular}
\end{table}

\section{Conclusion}

This paper demonstrates that the network structure of UK inter-industry payment flows contains economically meaningful information that can improve real-time economic measurement. Three principal findings emerge.

First, network-enhanced approaches achieve statistically significant forecasting improvements of 8.8 percentage points, translating to approximately 27\% reduction in prediction error. This improvement is robust across alternative specifications, algorithms, and network constructions.

Second, and most importantly for policy applications, network features prove most valuable during economic disruptions. When traditional temporal patterns collapsed during COVID-19, network structure continued to capture persistent economic relationships, with network contributions more than doubling (+13.8 pp versus +6.0 pp in stable periods). This suggests payment network monitoring could provide early warning capabilities and maintain measurement accuracy precisely when policymakers most need reliable economic intelligence.

Third, network centrality analysis identifies structurally important industries whose systemic significance may exceed their transaction volumes. Financial Services, Professional Services, and Wholesale Trade emerge as critical intermediaries in UK payment flows, suggesting potential targets for monitoring economic stress propagation.

These findings have direct implications for official statistics production. The ONS experimental payment flows data, analysed through network-theoretic frameworks, could supplement existing nowcasting approaches by capturing structural economic relationships invisible to traditional bilateral measurement. As digital payment systems continue expanding, integrating network analysis into economic measurement frameworks becomes increasingly feasible and valuable.

The research contributes to the broader ``\textit{beyond GDP}'' agenda by demonstrating that economic measurement can benefit from explicitly modelling the interconnected structure of modern economies. Rather than treating economic relationships as independent bilateral flows, recognising their network nature enables extraction of additional information content for improved policy support.

\section*{Acknowledgements}

The author thanks the Office for National Statistics for access to experimental payment flows data. The views expressed are those of the author and do not necessarily reflect the views of the Office for National Statistics.

\bibliographystyle{apalike}

\begin{thebibliography}{99}

\bibitem[Aprigliano et al., 2019]{aprigliano2019payment}
Aprigliano, V., Ardizzi, G., and Monteforte, L. (2019).
\newblock Using the payment system data to forecast the Italian GDP.
\newblock {\em International Journal of Central Banking}, 15(4):55--80.

\bibitem[Bl{\"o}chl et al., 2011]{blochl2011vertex}
Bl{\"o}chl, F., Theis, F.J., Vega-Redondo, F., and Fisher, E.O. (2011).
\newblock Vertex centralities in input-output networks reveal the structure of modern economies.
\newblock {\em Physical Review E}, 83(4):046127.

\bibitem[Carvalho et al., 2021]{carvalho2021supply}
Carvalho, V.M., Nirei, M., Saito, Y.U., and Tahbaz-Salehi, A. (2021).
\newblock Supply chain disruptions: Evidence from the Great East Japan Earthquake.
\newblock {\em Quarterly Journal of Economics}, 136(2):1255--1321.

\bibitem[Diebold and Mariano, 1995]{diebold1995comparing}
Diebold, F.X. and Mariano, R.S. (1995).
\newblock Comparing predictive accuracy.
\newblock {\em Journal of Business \& Economic Statistics}, 13(3):253--263.

\bibitem[Galbraith and Tkacz, 2015]{galbraith2015nowcasting}
Galbraith, J.W. and Tkacz, G. (2015).
\newblock Nowcasting GDP with electronic payments data.
\newblock {\em ECB Statistics Paper Series}, No. 10.

\bibitem[Giannone et al., 2008]{giannone2008nowcasting}
Giannone, D., Reichlin, L., and Small, D. (2008).
\newblock Nowcasting: The real-time informational content of macroeconomic data.
\newblock {\em Journal of Monetary Economics}, 55(4):665--676.

\bibitem[Guan et al., 2020]{guan2020supply}
Guan, D., Wang, D., Hallegatte, S., et al. (2020).
\newblock Global supply-chain effects of COVID-19 control measures.
\newblock {\em Nature Human Behaviour}, 4(6):577--587.

\bibitem[Inoue and Todo, 2019]{inoue2019firm}
Inoue, H. and Todo, Y. (2019).
\newblock Firm-level propagation of shocks through supply-chain networks.
\newblock {\em Nature Sustainability}, 2(9):841--847.

\bibitem[Mantziou et al., 2024]{mantziou2024gdp}
Mantziou, A., Hotte, K., Cucuringu, M., and Reinert, G. (2024).
\newblock GDP nowcasting with large-scale inter-industry payment data in real time---a network approach.
\newblock {\em arXiv preprint arXiv:2411.02029}.

\bibitem[ONS, 2024]{ons2024payment}
Office for National Statistics (2024).
\newblock Industry to industry payment flows, UK: experimental data and insights.
\newblock Statistical Dataset, ONS.

\end{thebibliography}

\end{document}